\documentclass[letterpaper]{article} 
\usepackage{aaai2026}  
\usepackage{times}  
\usepackage{helvet}  
\usepackage{courier}  
\usepackage[hyphens]{url}  
\usepackage{graphicx} 
\usepackage{natbib}  
\usepackage{caption} 
\usepackage{algorithm}
\usepackage{algorithmic}
\usepackage{amsmath}
\usepackage{booktabs}
\usepackage{multirow}
\usepackage{tikz}
\usepackage{tcolorbox}
\usepackage{newfloat}
\usepackage{listings}
\DeclareCaptionStyle{ruled}{labelfont=normalfont,labelsep=colon,strut=off} 
\floatstyle{ruled}
\newfloat{listing}{tb}{lst}{}
\floatname{listing}{Listing}
\nocopyright

\title{\textsc{HiPlan}: Hierarchical Planning for LLM Agents\\with Adaptive Global-Local Guidance}

\author{
    Ziyue Li\textsuperscript{\rm 1,\rm 2},
    Yuan Chang\textsuperscript{\rm 1,\rm 2},
    Gaihong Yu\textsuperscript{\rm 1}\thanks{Corresponding Author},
    Xiaoqiu Le\textsuperscript{\rm 1,\rm 2}\footnotemark[1]
}

\affiliations{
     \textsuperscript{1}National Science Library, Chinese Academy of Sciences \\
 \textsuperscript{2}Department of Information Resources Management, School of Economics and Management, \\
 University of Chinese Academy of Sciences 

    \{liziyue, changyuan, yugh, lexq\}@mail.las.ac.cn
}

\usepackage{bibentry}

\begin{document}

\maketitle

\begin{abstract}
Large language model (LLM)-based agents have demonstrated remarkable capabilities in decision-making tasks, but struggle significantly with complex, long-horizon planning scenarios. This arises from their lack of macroscopic guidance, causing disorientation and failures in complex tasks, as well as insufficient continuous oversight during execution, rendering them unresponsive to environmental changes and prone to deviations. To tackle these challenges, we introduce \textsc{HiPlan}, a hierarchical planning framework that provides adaptive global-local guidance to boost LLM-based agents’ decision-making. \textsc{HiPlan} decomposes complex tasks into milestone action guides for general direction and step-wise hints for detailed actions. During the offline phase, we construct a milestone library from expert demonstrations, enabling structured experience reuse by retrieving semantically similar tasks and milestones. In the execution phase, trajectory segments from past milestones are dynamically adapted to generate step-wise hints that align current observations with the milestone objectives, bridging gaps and correcting deviations. Extensive experiments across two challenging benchmarks demonstrate that \textsc{HiPlan} substantially outperforms strong baselines, and ablation studies validate the complementary benefits of its hierarchical components.
\end{abstract}


\section{Introduction}

Large language model (LLM)-based agents have recently demonstrated remarkable capabilities in a wide range of decision-making and reasoning tasks\cite{durante2024agentaisurveyinghorizons,xi2025rise,wang2024survey,huang2024understandingplanningllmagents}. Their proficiency in understanding complex instructions and generating coherent, context-aware responses has enabled the automation of various applications. Despite these advances, enabling LLM-based agents to effectively plan and act over long horizons remains a fundamental challenge, especially given the complexity of real-world tasks and the dynamic nature of environments.

\begin{figure}[t]
    \centering
    \includegraphics[width=\columnwidth]{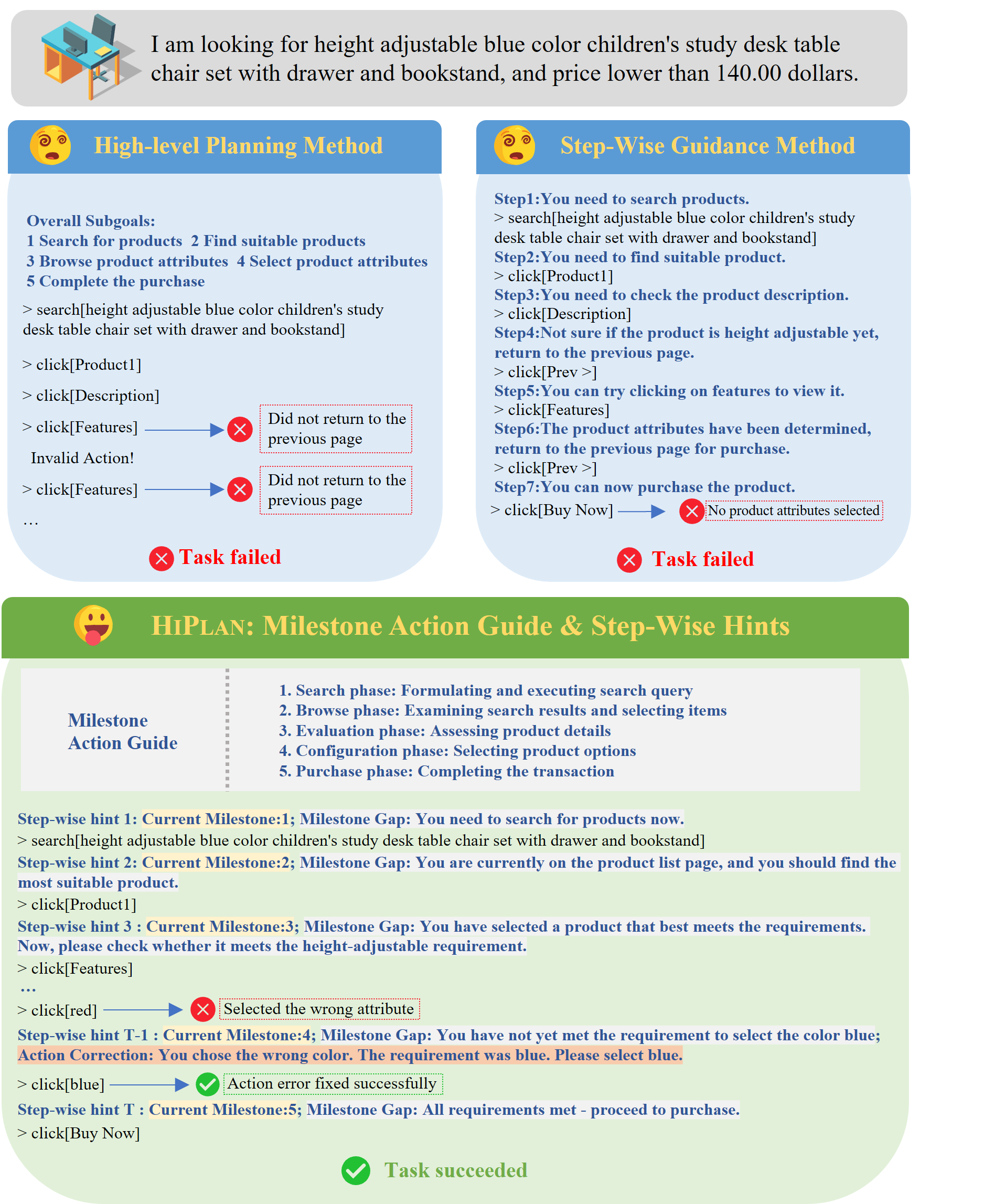}
    \caption{\textbf{Top-Left}: High-level planning with global subgoals lacking flexibility; \textbf{Top-Right}: Step-wise methods with local adaptability but limited global guidance; \textbf{Bottom}: \textsc{HiPlan}’s hierarchical approach combining milestone guidance and step-wise hints for adaptive and robust planning. }
    \label{fig:introduction}
\end{figure}

Existing research has approached these challenges from two main perspectives. High-level planning methods \cite{khot2023decomposed,wang-etal-2023-plan} allow agents to decompose tasks into subgoals, providing clear overall direction and guiding the agent’s progress toward fulfilling the final objectives. However, such methods often exhibit limited flexibility in handling unexpected execution errors or dynamically adapting actions when the environment changes (see Fig.\ref{fig:introduction}; Top-Left). Furthermore, these methods often rely on complete historical trajectories of specific tasks as demonstrations, which introduces excessive task-specific details that hinder generalization and reduce robustness in new scenarios. Step-wise methods \cite{yao2023react,DBLP:conf/sigir/ZhouYW0WXXY024} excel at adapting actions to real-time observations. However, this finer-grained focus frequently leads the agent to lose sight of the overall task structure, making it prone to inefficient or locally optimal behaviors—especially in long-horizon or unfamiliar tasks (see Fig.\ref{fig:introduction}; Top-Right). Moreover, step-wise methods often struggle to effectively reuse prior experience beyond immediate observations, hindering their scalability and adaptability in diverse tasks.

To overcome existing limitations, we propose \textsc{HiPlan}, a hierarchical planning framework that endows LLM-based agents with adaptive global-local guidance (see Fig.\ref{fig:introduction}; Bottom). At the macro level, \textsc{HiPlan} employs a milestone action guide as a 'roadmap' delineating critical task stages to maintain global direction and avoid local optima, while at the micro level, step-wise hints act like real-time 'traffic updates', providing fine-grained feedback to correct actions and align progress with current milestones. This synergy enhances efficiency, controllability, and overall robustness.

A core innovation of \textsc{HiPlan} is its effective reuse of historical experience through a milestone library constructed offline from expert demonstrations. During execution, this library guides the generation of the high-level milestone action guide, enabling the agent to learn from prior experience at a macro scale. For the step-wise hint generation, \textsc{HiPlan} retrieves trajectory fragments corresponding to similar completed milestones, providing fine-grained, context-relevant guidance. We select milestone-level experience for reuse because action-level trajectories are often too dependent on specific contexts to offer useful information, while task-level trajectories include excessive details that introduce noise. Positioned between these two, milestone-level trajectories serve as ideal units of intermediate granularity, balancing informativeness and generalizability for effective retrieval and planning.

We evaluate \textsc{HiPlan} on two challenging benchmarks, ALFWorld \cite{shridhar2021alfworld} and WebShop \cite{NEURIPS2022_82ad13ec}, featuring complex long-horizon tasks. Results show \textsc{HiPlan} consistently outperforms strong baselines with higher success rates and greater robustness. Ablation studies confirm the essential contributions of milestone action guide and step-wise hints to the overall performance.
Our main contributions are summarized as follows:

\begin{itemize}
    \item We introduce \textsc{HiPlan}, a novel hierarchical planning framework that tightly integrates global milestone action guides with local step-wise hints, achieving adaptive global-local guidance for agent planning.
    \item We propose an efficient milestone-level experience reuse strategy that allows agents to draw on prior demonstrations in a way that is both generalizable and actionable.
    \item We conduct extensive experiments on multiple challenging benchmarks, demonstrating that \textsc{HiPlan} significantly improves task success rates and robustness compared to strong baselines, confirming its effectiveness across diverse decision-making scenarios.
\end{itemize}

\section{Related Work}
\subsection{LLM-Based Agent for Planning}
LLMs have demonstrated remarkable capabilities across diverse domains, excelling in various aspects such as complex reasoning \cite{wei2022chain,yao2023tree}, problem-solving \cite{li-etal-2024-simulating,xu-etal-2024-chatglm}, and text generation\cite{gao-etal-2023-enabling,chang2025treereviewdynamictreequestions}. Building on these core competencies, recent advancements have extended their application to more challenging scenarios, where LLM-based agents exhibit promising potential in autonomous decision-making and planning.

Current planning approaches for LLM-based agents can be broadly categorized into two primary paradigms. High-level planning methods \cite{erdogan2025planandact,sun-etal-2024-pearl,wang-etal-2023-plan,khot2023decomposed} focus on decomposing complex tasks into structured subgoals or generating comprehensive plans before execution. These approaches provide clear overall direction and maintain global coherence throughout task execution. However, such methods often exhibit limited flexibility when encountering unexpected execution errors or adapting to dynamic environmental changes. Step-wise planning methods \cite{yao2023react,DBLP:conf/sigir/ZhouYW0WXXY024,nguyen2024steptimelanguageagents} excel at real-time adaptation by interleaving reasoning and action steps. These methods enable agents to adjust their strategies based on immediate observations and environmental feedback, making them highly responsive to changing conditions. Nevertheless, this fine-grained focus frequently leads agents to lose sight of the overall task structure, resulting in inefficient exploration or locally optimal behaviors, particularly in long-horizon scenarios.

Additional representative approaches encompass memory-augmented systems \cite{zhao2024expel,hu2024hiagenthierarchicalworkingmemory} that leverage historical experiences for improved decision-making, and reflection-based frameworks \cite{shinn2023reflexion} that enable agents to learn from failures through self-critique and iterative improvement.

Despite these advances, existing methods face fundamental limitations in achieving both global coherence and local adaptability simultaneously. Our work addresses this by integrating global milestone action guides with local step-wise hints, enabling adaptive global-local guidance for enhanced efficiency and robustness in long-horizon tasks.

\subsection{Retrieval-Augmented Planning}
Retrieval-augmented planning (RAP) methods enhance LLM‑based agents by retrieving past experiences—trajectories, plans, or instruction graphs—to ground planning in real execution data. 

One approach retrieves relevant exemplars or context fragments conditioned on task similarity, enabling improved planning in both text-only and multimodal environments \cite{liu-etal-2022-makes,trivedi-etal-2023-interleaving,DBLP:conf/sigir/ZhouYW0WXXY024}. A second strand organizes retrieval around abstract, structured representations—such as instruction graphs—to improve transferability and generalization \cite{kim-etal-2024-rada,wang2025instructragleveragingretrievalaugmentedgeneration}.

Despite their strength, existing RAP approaches often rely heavily on full exemplar retrieval, which can introduce noise and limit flexibility. They also typically decouple global planning from local adaptability. In contrast, \textsc{HiPlan} integrates task-level and milestone-level retrieval, which enables robust, structured guidance that retains adaptability to execution context—addressing both generalization and dynamic control more effectively than prior RAP methods.

\section{Preliminaries}

We study the problem of long-horizon task completion in partially observable environments, where an agent must generate a coherent sequence of actions to fulfill a natural language instruction. The environment is formalized as a partially observable Markov decision process (POMDP), represented as \( \mathcal{M} = (\mathcal{S}, \mathcal{A}, \mathcal{O}, T) \), where \( \mathcal{S} \) denotes the set of latent environment states, \( \mathcal{A} \) the discrete action space, \( \mathcal{O} \) the observation space, and \( T: \mathcal{S} \times \mathcal{A} \rightarrow \mathcal{S} \) the transition function.

At the beginning of each episode, the agent is provided with a task instruction \( \tau \in \mathcal{T} \), expressed in natural language. Over a sequence of steps \( t = 1, 2, \ldots, T \), the agent receives observations \( o_t \in \mathcal{O} \) and selects actions \( a_t \in \mathcal{A} \), aiming to reach a successful terminal state that satisfies \( \tau \), despite partial observability and complex dynamics.

\section{Method}
\begin{figure*}[t]
\centering
\includegraphics[width=\textwidth]{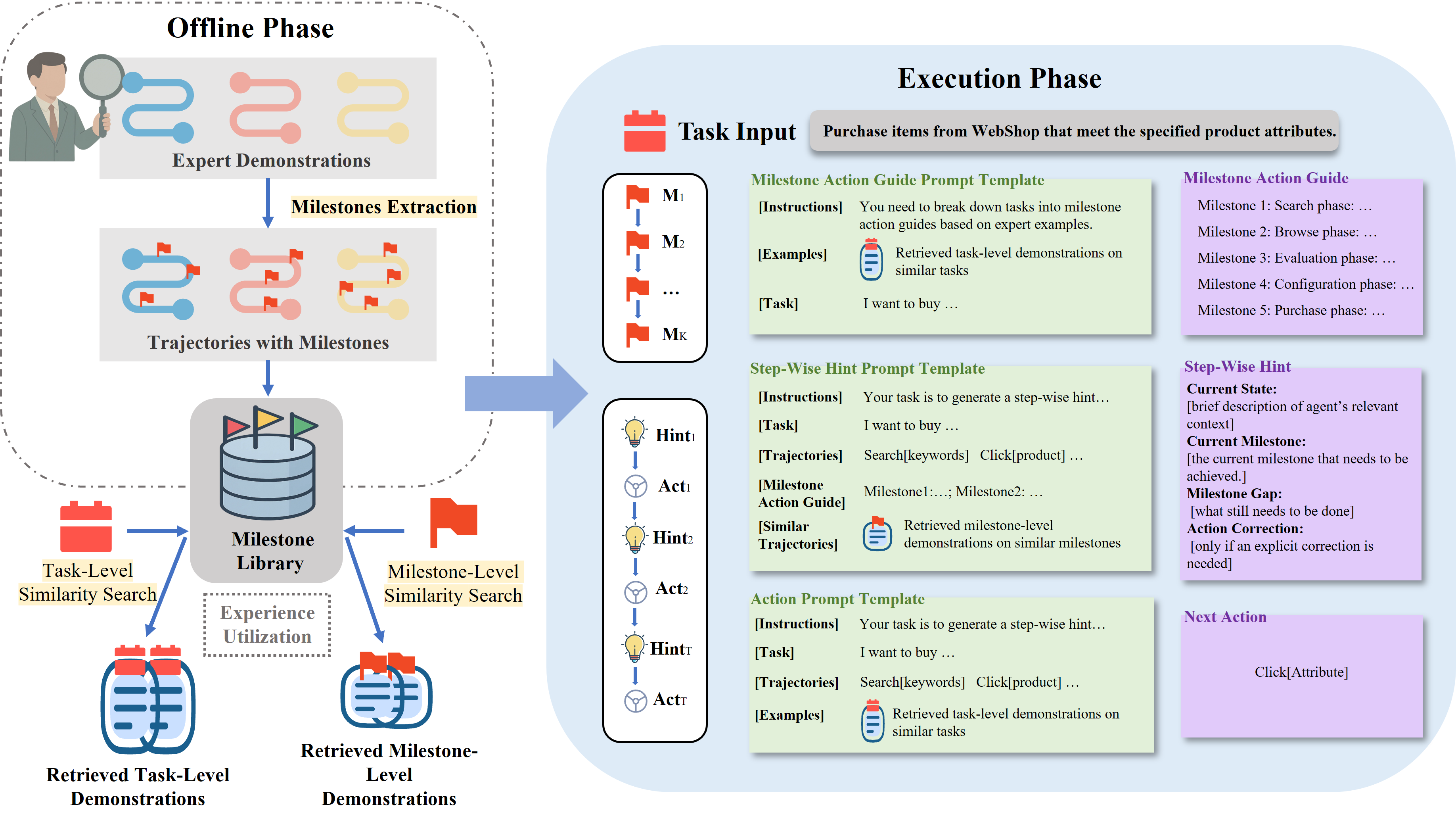}
\caption{The \textsc{HiPlan} framework. In the offline phase (top left), a milestone library is constructed from expert demonstrations. During online execution (right), the agent utilizes this library by retrieving relevant task and milestone-level experiences to generate a global Milestone Action Guide and local Step-Wise Hints, enabling adaptive planning.}
\label{fig:framework}
\end{figure*}
We propose \textsc{HiPlan}, a hierarchical planning framework that equips LLM-based agents with adaptive global-local guidance for tackling long-horizon tasks. 

\subsection{Overview}
Fig. \ref{fig:framework} illustrates the overall architecture and workflow of \textsc{HiPlan}. Our method decomposes complex tasks into a sequence of critical milestones forming a high-level action guide, while progressively generating fine-grained step-wise hints to refine each action based on real-time observations. 

To facilitate structured planning and control, we assume access to a set of successful task demonstrations \( \mathcal{D} = \{ (\tau^{(i)}, \xi^{(i)}) \}_{i=1}^N \), where each trajectory \( \xi^{(i)} = \left[ (o_1^{(i)}, a_1^{(i)}), \ldots, (o_{T^{(i)}}^{(i)}, a_{T^{(i)}}^{(i)}) \right] \) corresponds to a task \( \tau^{(i)} \). From these demonstrations, \textsc{HiPlan} constructs a milestone library and retrieves structured experience during execution.

We introduce two complementary forms of guidance for hierarchical planning:

\begin{itemize}
    \item \textbf{Global Guidance:} A milestone action guide \( \mathcal{G}_\tau = [m_1, m_2, \ldots, m_K] \), where each \( m_k \) is a natural language subgoal that represents a critical stage in completing the task. This sequence provides a high-level plan structure which serves as the coarse-grained directional guidance.

    \item \textbf{Local Guidance:} A step-wise hint \( h_t \), generated at each timestep by retrieving trajectory fragments that correspond to the current milestone. This hint offers fine-grained behavioral suggestions conditioned on the current observation and subgoal.
\end{itemize}

The agent's policy \( \pi: (\tau, o_t, \mathcal{G}_\tau, h_t) \rightarrow a_t \) integrates both levels of guidance to produce context-aware, goal-directed actions. The combination of a global roadmap and localized hint and retrieval enables robust and efficient decision-making in complex long-horizon tasks.

\subsection{Offline Phase: Milestone Library Construction}

\textsc{HiPlan} constructs a milestone library \( \mathcal{M}_L \) from the set of successful demonstrations \( \mathcal{D} = \{ (\tau^{(i)}, \xi^{(i)}) \}_{i=1}^N \). For each trajectory \( \mathcal{\xi^{(i)}} \), we segment it into \( K^{(i)} \) contiguous fragments \( \{ \zeta_k^{(i)} \}_{k=1}^{K^{(i)}} \), each corresponding to a semantically meaningful subgoal. An LLM is prompted to generate a natural language description \( m_k^{(i)} \) for each segment, forming a milestone sequence \( \mathcal{G}_{\tau^{(i)}} = [m_1^{(i)}, \ldots, m_{K^{(i)}}^{(i)}] \).

To support efficient retrieval, each instruction and milestone is embedded into a dense vector space. The milestone library stores tuples of the form:
\[
(\mathbf{v}_{\text{task}}^{(i)}, \mathbf{v}_{\text{milestone}}^{(i,k)}, \tau^{(i)}, m_k^{(i)}, \zeta_k^{(i)}),
\]
where \( \mathbf{v}_{\text{task}}^{(i)} \) and \( \mathbf{v}_{\text{milestone}}^{(i,k)} \) are vector representations for the task \( \tau^{(i)} \) and the milestone \( m_k^{(i)} \), respectively. Similarity is computed via dot product over normalized embeddings.

The final milestone library is defined as:
\[
\mathcal{M}_L = \bigcup_{i=1}^N \bigcup_{k=1}^{K^{(i)}} \left\{ (\mathbf{v}_{\text{task}}^{(i)}, \mathbf{v}_{\text{milestone}}^{(i,k)}, m_k^{(i)}, \zeta_k^{(i)}) \right\}.
\]
The milestone library abstracts higher-level structured experiences while preserving task-specific low-level details. Compared to raw trajectory or task-level retrieval, this mid-granularity representation provides a balanced trade-off between generalizability and specificity, making it ideal for hierarchical planning under varied conditions.

\subsection{Execution Phase: Hierarchical Planning and Execution}

In the execution phase, \textsc{HiPlan} performs hierarchical planning by dynamically integrating global milestone guidance and fine-grained step-wise hints. Leveraging structured experiences from the milestone library, our framework provides an adaptive planning mechanism that maintains global consistency and facilitates local flexibility, as illustrated in Fig.~\ref{fig:framework}. We assume this dual-level approach can help mitigate common issues in long-horizon task execution, such as deviation from global objectives and inability to adapt to real-time uncertainties.

\paragraph{Global Guidance: Milestone Action Guide} To establish strategic direction for long-horizon task completion, \textsc{HiPlan} provides global guidance through a dynamically generated milestone action guide. 
Given a test-time task instruction \(\tau\), \textsc{HiPlan} retrieves similar task entries from the milestone library using the embedding of \(\tau\) as the query key:
\begin{equation}
    \{ (\tau^{(j)}, \xi^{(j)}, \mathcal{G}_{\tau^{(j)}}) \}_{j=1}^M = \text{Retrieve}(\mathbf{v}_\tau)
\end{equation}

These entries include task instructions, trajectories, and corresponding milestone sequences, which serve as references to generate a tailored guide:
\begin{equation}
    \mathcal{G}_\tau = \text{LLM}(\tau, \{ (\tau^{(j)}, \xi^{(j)}, \mathcal{G}_{\tau^{(j)}}) \}_{j=1}^M),
\end{equation}
where \(\mathcal{G}_\tau = [m_1, m_2, \ldots, m_K]\) and each \(m_k\) is a critical subgoal adapted to the current task context. 

This process aims to transfer insights from historical experiences, enabling the agent to benefit from past successful planning strategies while maintaining flexibility to handle new task dynamics.

\paragraph{Local Guidance: Step-Wise Hints}
Complementing the global milestone structure, \textsc{HiPlan} provides detailed local guidance through context-awareness step-wise hints, dynamically generated at each step. At timestep \(t\), the current milestone \(m_{\psi(t)}\) is identified, and its embedding \(\mathbf{v}_{m_{\psi(t)}}\) serves as the query key to retrieve similar milestones and their corresponding trajectory segments from the milestone library:
\begin{equation}
    \{ (m_l^*, \zeta_l^*) \}_{l=1}^P = \text{Retrieve}(\mathbf{v}_{m_{\psi(t)}}),
\end{equation}
where \(m_l^*\) and \(\zeta_l^*\) denote retrieved milestones and the trajectory segments that complete them, respectively. These retrieved elements are then used as references, combined with past action-observation pairs  \(\{(o_s, a_s)\}_{s=1}^t \), to generate the step-wise hint \(h_t\):
\begin{equation}
    h_t = \text{LLM}(m_k, \{(o_s, a_s)\}_{s=1}^t, \{ (m_l^*, \zeta_l^*) \}_{l=1}^P).
\end{equation}

 Each hint explicitly highlights the current state context, the gap to the milestone, and, when necessary, corrections to the agent’s intended actions, thus providing immediate feedback to rectify errors or inefficiencies:
\[
h_t = \{ \text{State Context}, \text{Milestone Gap}, \text{Action Correction}\}.
\]

The step-wise hints dynamically track the agent's progression, recognizing when milestone subgoals are achieved and seamlessly guiding the agent toward subsequent objectives.

\paragraph{Dual-level Guidance Enhanced Policy} At each timestep \(t\), the agent leverages the milestone action guide to maintain global task coherence, while simultaneously utilizing step-wise hints to adaptively transition between milestones based on real-time observations.

Formally, the integrated policy \(\pi\) combines these two complementary sources of guidance as:
\begin{equation}
    a_t = \pi(\tau, \{(o_s, a_s)\}_{s=1}^t, m_k, h_t)
\end{equation}

In this formulation, \textsc{HiPlan} performs adaptive hierarchical planning by closely integrating the global milestone action guide with dynamic local step-wise hints. The detailed algorithmic workflow is presented in Alg.~\ref{alg:hiplan}.

\begin{algorithm}[tb]
\caption{\textsc{HiPlan}: Hierarchical Planning with Adaptive Global-Local Guidance}
\label{alg:hiplan}
\textbf{Input}: Task instruction $\tau$ \\
\textbf{Parameter}: Demonstration set $\mathcal{D} = \{ (\tau^{(i)}, \xi^{(i)}) \}_{i=1}^N$ \\
\textbf{Output}: Actions $[a_1, a_2, \dots, a_T]$
\begin{algorithmic}[1]
\STATE \textbf{// Offline Phase: Milestone Library Construction}
\FOR{$(\tau^{(i)}, \xi^{(i)}) \in \mathcal{D}$}
    \FOR{each fragment $\zeta_k^{(i)}$ in $\xi^{(i)}$}
        \STATE $m_k^{(i)} \leftarrow \text{LLM}(\zeta_k^{(i)})$
        \STATE $(\mathbf{v}_{\text{task}}^{(i)}, \mathbf{v}_{\text{mile}}^{(i,k)}) \leftarrow \text{Embed}(\tau^{(i)}, m_k^{(i)})$
        \STATE Store $(\mathbf{v}_{\text{task}}^{(i)}, \mathbf{v}_{\text{mile}}^{(i,k)}, m_k^{(i)}, \zeta_k^{(i)})$ in $\mathcal{M}_L$
    \ENDFOR
\ENDFOR

\STATE \textbf{// Execution Phase: Hierarchical Planning}
\STATE $\mathbf{v}_\tau \leftarrow \text{Embed}(\tau)$
\STATE Retrieve $\{(\tau^{(j)}, \xi^{(j)}, \mathcal{G}_{\tau^{(j)}})\}_{j=1}^M$ from $\mathcal{M}_L$ using $\mathbf{v}_\tau$
\STATE $\mathcal{G}_\tau \leftarrow \text{LLM}(\tau, \{(\tau^{(j)}, \xi^{(j)}, \mathcal{G}_{\tau^{(j)}})\})$
\STATE $k \leftarrow 1$ \COMMENT{Milestone index}

\FOR{$t = 1$ to $T$}
    \STATE $o_t \leftarrow$ observe environment
    \STATE Retrieve $\{(m_l^*, \zeta_l^*)\}_{l=1}^P$ from $\mathcal{M}_L$ using $\mathbf{v}_{m_k}$
    \STATE $h_t \leftarrow \text{LLM}(m_k, \{(o_s, a_s)\}_{s=1}^t, \{(m_l^*, \zeta_l^*)\})$
    \STATE $a_t \leftarrow \pi(\tau, \{(o_s, a_s)\}_{s=1}^t, m_k, h_t)$
    \STATE Execute $a_t$
    \IF{milestone $m_k$ completed}
        \STATE $k \leftarrow k + 1$
    \ENDIF
    \IF{task completed or $t \geq T$}
        \STATE \textbf{break}
    \ENDIF
\ENDFOR

\STATE \textbf{return} $[a_1, a_2, \dots, a_T]$
\end{algorithmic}
\end{algorithm}

\section{Experiment}
\subsection{Experimental Setup}

\subsubsection{Datasets}
We evaluate \textsc{HiPlan} on two widely used benchmarks for long-horizon decision-making: 

\textbf{ALFWorld} \cite{shridhar2021alfworld} is a text-based benchmark that challenges an agent's ability to perform complex, multi-step tasks in a simulated household environment. The benchmark comprises six distinct task types: Pick \& Place, Pick Two \& Place, Examine in Light, Clean \& Place, Heat \& Place, and Cool \& Place, testing an agent's capacity for understanding object states and interactions, and executing long action sequences that can exceed 50 steps. Our evaluation is conducted on the established split of 134 out-of-distribution tasks.

\textbf{WebShop} \cite{NEURIPS2022_82ad13ec} is a large-scale interactive environment that simulates an online shopping website with over 1.18 million products. It tests an agent's ability to ground natural language instructions into a sequence of search and click actions to purchase a specific product. We evaluate \textsc{HiPlan} on a set of 200 test instructions, reporting the average reward, which reflects the degree of attribute alignment with the request, and the success rate, which measures the fraction of tasks where all specifications are satisfied perfectly.

\subsubsection{Implementation Details}

To construct the milestone library, we collect successful expert trajectories from both ALFWorld and WebShop. Each trajectory is segmented into key milestones using GPT-4o (\texttt{gpt-4o-2024-08-06}). Task instructions and extracted milestone descriptions are encoded using the SentenceTransformers model \texttt{all-mpnet-base-v2}, and indexed via inner-product similarity for efficient retrieval.

All components of \textsc{HiPlan}—including milestone action guide generation, step-wise hint construction, and final action prediction—are executed by the same underlying LLM. We evaluate the framework using two open-source models: \texttt{Mixtral-8x22B-Instruct-v0.1} (denoted as Mixtral) and \texttt{LLaMA-3.3-70B-Instruct} (denoted as LLaMA). These two represent distinct model architectures: Mixtral is a sparse mixture-of-experts (SMoE) model, while LLaMA is a dense pre-trained LLM. This choice allows us to assess \textsc{HiPlan}’s generality across different model types while ensuring reproducibility.

In line with prior work, we limit each episode to at most 50 steps in ALFWorld and 40 steps in WebShop. All outputs from LLMs are generated deterministically with temperature set to 0.0 to ensure consistent and reproducible results.

\subsubsection{Baselines}
We compare \textsc{HiPlan} against three strong LLM-based planning baselines, each representative of a distinct approach to decision-making:

\textbf{REACT} \cite{yao2023react} exemplifies \emph{classical planning-style methods}, where reasoning and acting are interleaved through chain-of-thought prompting and action generation. It enables the agent to maintain high-level plans, update them on the fly, and interface with the environment in a structured manner.

\textbf{Reflexion} \cite{shinn2023reflexion} represents the class of \emph{reflection-driven methods}. Instead of parameter updates, it leverages self-generated linguistic feedback to iteratively improve performance across episodes. The agent reflects on past outcomes, maintains episodic memory, and refines its behavior through natural language reasoning. 

\textbf{TRAD} \cite{DBLP:conf/sigir/ZhouYW0WXXY024} is a \emph{retrieval-augmented method} that improves in-context decision-making by selecting trajectory segments based on thought-level similarity. By aligning these fragments and filtering irrelevant context, TRAD enables agents to draw more directly on prior demonstrations. 

All baselines are evaluated under the same conditions as \textsc{HiPlan}, using identical environments and LLM backbones to ensure a fair comparison. Implementation details for all baselines are provided in Appendix.

\subsection{Main Results}

\begin{table*}[t]
\centering
\renewcommand{\arraystretch}{1.2}
\resizebox{\textwidth}{!}{
\begin{tabular}{llccccccc}
\toprule
\textbf{Models} & \textbf{Method} & \textbf{Put} & \textbf{Examine} & \textbf{Clean} & \textbf{Heat} & \textbf{Cool} & \textbf{PutTwo} & \textbf{All} \\
\midrule
\multirow{4}{*}{\textbf{Mixtral-8x22b}} 
    & REACT              & 0.46 & 0.89 & 0.55 & 0.83 & 0.62 & 0.18 & 0.59 \\
    & REACT+Reflexion    & 0.77 & 0.89 & 0.61 & 0.83 & 0.62 & 0.29 & 0.64 \\
    & TRAD               & 0.75 & 0.89 & \textbf{0.84} & 0.70 & \textbf{0.95} & 0.53 & 0.78 \\
    & \textbf{\textsc{HiPlan}}               & \textbf{0.92} & \textbf{0.94} & \textbf{0.84} & \textbf{0.87} & 0.71 & \textbf{0.59} & \textbf{0.82} \\
\midrule
\multirow{4}{*}{\textbf{LLaMA3.3-70b}} 
    & REACT              & 0.50 & 0.83 & 0.39 & 0.65 & 0.48 & 0.18 & 0.50 \\
    & REACT+Reflexion    & 0.58 & 0.83 & 0.48 & 0.65 & 0.57 & 0.24 & 0.56 \\
    & TRAD               & 0.96 & 0.89 & 0.90 & 0.65 & 0.67 & 0.59 & 0.79 \\
    & \textbf{\textsc{HiPlan}}               & \textbf{1.00} & \textbf{1.00} & \textbf{0.97} & \textbf{0.91} & \textbf{0.90} & \textbf{0.82} & \textbf{0.94} \\
\bottomrule
\end{tabular}
}
\caption{Task success rates of \textsc{HiPlan} and baselines on the ALFWorld benchmark across six task types.}
\label{tab:alfworld_result}
\end{table*}

\begin{table}[ht]
\centering
\renewcommand{\arraystretch}{1.15}
\setlength{\tabcolsep}{4pt}
\begin{tabular}{@{}p{2.3cm} p{2.2cm} c c@{}}
\toprule
\textbf{Models} & \textbf{Method} & \textbf{Reward} & \textbf{Success Rate} \\
\midrule
\multirow{4}{=}{\textbf{Mixtral-8x22b}} 
    & REACT             & 0.26 & 0.19 \\
    & REACT+Reflexion   & 0.40 & 0.24 \\
    & TRAD              & 0.10 & 0.04 \\
    & \textbf{\textsc{HiPlan}}              & \textbf{0.50} & \textbf{0.36} \\
\midrule
\multirow{4}{=}{\textbf{LLaMA3.3-70b}} 
    & REACT             & 0.09 & 0.08 \\
    & REACT+Reflexion   & 0.23 & 0.12 \\
    & TRAD              & 0.27 & 0.14 \\
    & \textbf{\textsc{HiPlan}}              & \textbf{0.58} & \textbf{0.40} \\
\bottomrule
\end{tabular}
\caption{Average reward and success rate of \textsc{HiPlan} and baselines on the WebShop benchmark.}
\label{tab:webshop_result}
\end{table}

We summarize our experimental results in Tables~\ref{tab:alfworld_result} and~\ref{tab:webshop_result}, showing that \textsc{HiPlan} consistently outperforms all baseline methods across both ALFWorld and WebShop benchmarks. We discuss these results in detail below.

\subsubsection{ALFWorld} \textsc{HiPlan} achieves the highest success rates across all task categories, demonstrating substantial improvements over baselines with absolute gains ranging from 4\% to 23\% points for Mixtral and 15\% to 44\% points for LLaMA. While Reflexion method enables agent to reflect on failed trials and retry, its reliance on episode-level feedback makes it struggle to escape similar error patterns, resulting in only marginal improvements over REACT. \textsc{HiPlan} also surpasses the strong retrieval-based TRAD baseline, which only supplies action-level demonstrations without higher-level goal structure and can introduce noise that misleads the agent. Notably, \textsc{HiPlan} demonstrates larger improvements on challenging tasks like \textit{PutTwo}, which involves multiple target objects and requires longer action sequences, indicating that hierarchical guidance becomes increasingly valuable as task complexity rises. 

\subsubsection{WebShop} Results shown in Table~\ref{tab:webshop_result} reveals a similar trend, with \textsc{HiPlan} achieving the highest success rates of 36\% (Mixtral) and 40\% (LLaMA), outperforming baselines by up to 32\% absolute points. All baseline methods perform poorly on this environment. Interestingly, TRAD shows inconsistent performance, even underperforming REACT and Reflexion on the Mixtral. This suggests that in WebShop environment, directly providing low-level action demonstrations as in-context examples can often introduce significant noise, severely misguiding the agent and preventing target product purchase within limited steps. This also indicating the sensitivity of TRAD to different task characteristics. \textsc{HiPlan} achieves consistent and substantial improvements in both average task reward (measuring attribute alignment of purchased products) and overall success rate. The superior task reward indicates that even when \textsc{HiPlan} fails to find the exact target product, it can identifies alternative products that better satisfy the specified constraints, guided by the synergy between high-level and step-wise signals.

Overall, these experimental results highlight the effectiveness of \textsc{HiPlan}’s hierarchical guidance mechanism. By leveraging structured milestone-level plans and providing dynamic global-local integration, \textsc{HiPlan} significantly enhances the performance of LLM-based agents in complex, long-horizon tasks. Its robust performance across different environments and model architectures underscores the generality and adaptability of our approach.

\subsection{Ablation Studies}
To assess the contribution of the core mechanisms in \textsc{HiPlan}, we conduct ablation studies across two environments by evaluating three variants: 

\begin{itemize}
\item \textbf{\textsc{HiPlan}-Direct:} Removes both the milestone action guide and step-wise hints, relying solely on direct action generation from task instructions, task-level similar demonstrations and observations.
\item \textbf{\textsc{HiPlan}-Milestone:} Retains only the high-level milestone action guide while removing the step-wise hints.
\item \textbf{\textsc{HiPlan}-w/o milestone-level demonstrations:} Includes both milestone action guides and step-wise hints, but constructs step-wise hints without milestone-level similar trajectory fragments as references from the milestone library.
\end{itemize}

\begin{figure}[t]
    \centering
\includegraphics[width=\columnwidth]{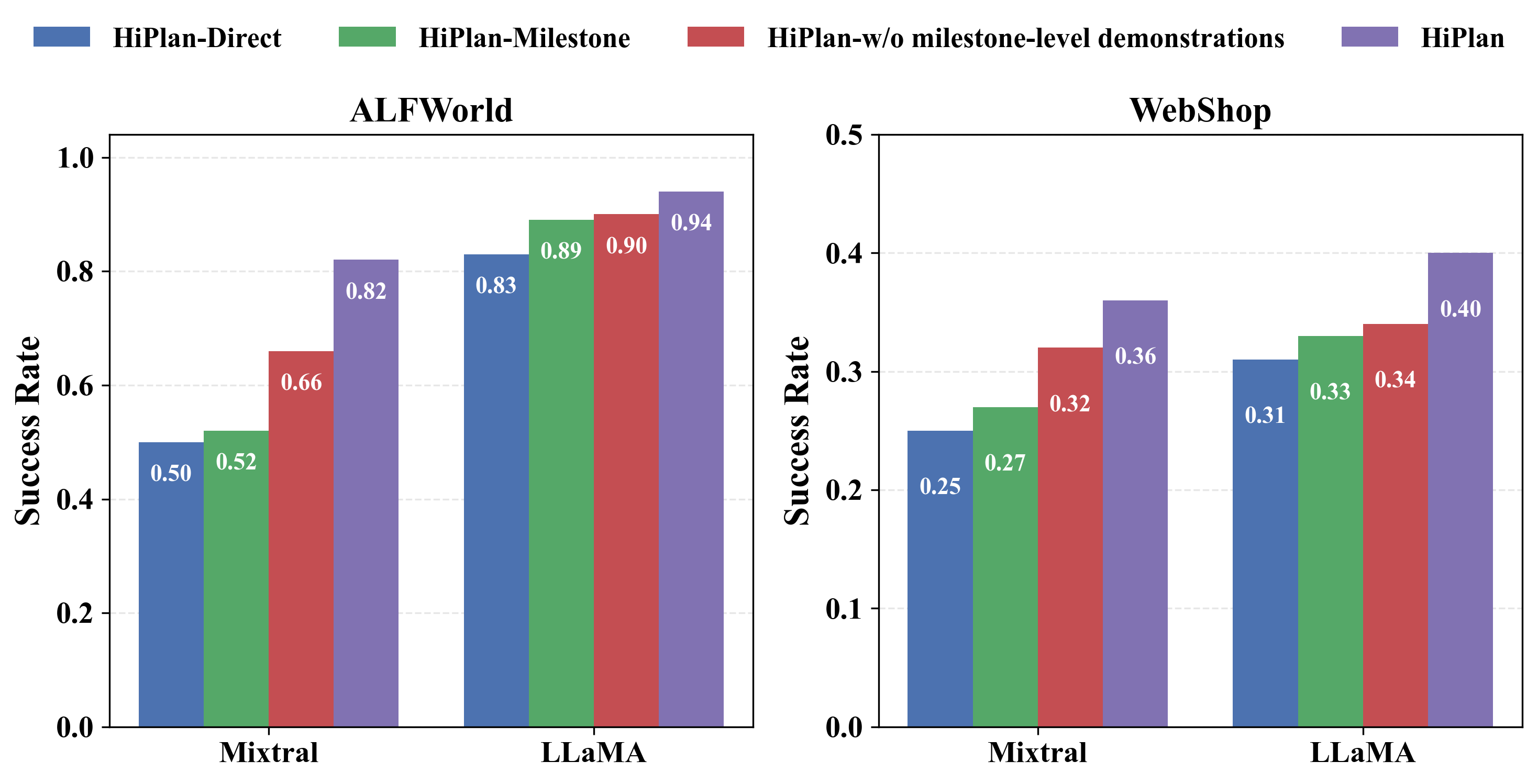}
    \caption{Ablation study results across ALFWorld and WebShop with Mixtral and LLaMA. \textsc{HiPlan} consistently outperforms all ablated variants, validating the importance of hierarchical guidance and the reuse of milestone-level demonstrations.}
    \label{fig:ablation}
\end{figure}

The results in Fig.~\ref{fig:ablation} demonstrate that all variants underperform \textsc{HiPlan} across environments and LLMs, which highlight several key findings:

\textbf{Hierarchical Guidance is Crucial.}  While \textsc{HiPlan}-Milestone provides modest improvements over the direct setting, the combination with step-wise hints yields substantial gains (11-32\% points in ALFWorld, 9-11\% points in WebShop). These results validate the synergy effect of \textsc{HiPlan}’s dual-level guidance.

\textbf{Milestone Demonstrations Enhance Planning Quality.}  \textsc{HiPlan}-w/o milestone-level demonstrations shows consistent performance drops. This validates the effect of leveraging milestone-level similar experiences from the milestone library rather than providing step-wise hints from scratch.

\textbf{Robust Generalizability. } The performance gains from \textsc{HiPlan} are consistent across both ALFWorld and WebShop, despite their distinct characteristics—text-based household versus web navigation. Both LLMs exhibit consistent cumulative effects from the components. This suggests that the benefits of hierarchical guidance and experiences reuse generalize across different scenarios and LLMs.

\subsection{Case Study}

To better understand how \textsc{HiPlan} enables adaptive hierarchical planning, we present a case study on the ALFWorld task: \textit{``put two soapbar in garbagecan''}. This task involves locating and manipulating multiple objects sequentially in a partially observable environment, making it a challenging scenario in this environment.

\begin{figure}[t]
    \centering
    \includegraphics[width=\columnwidth]{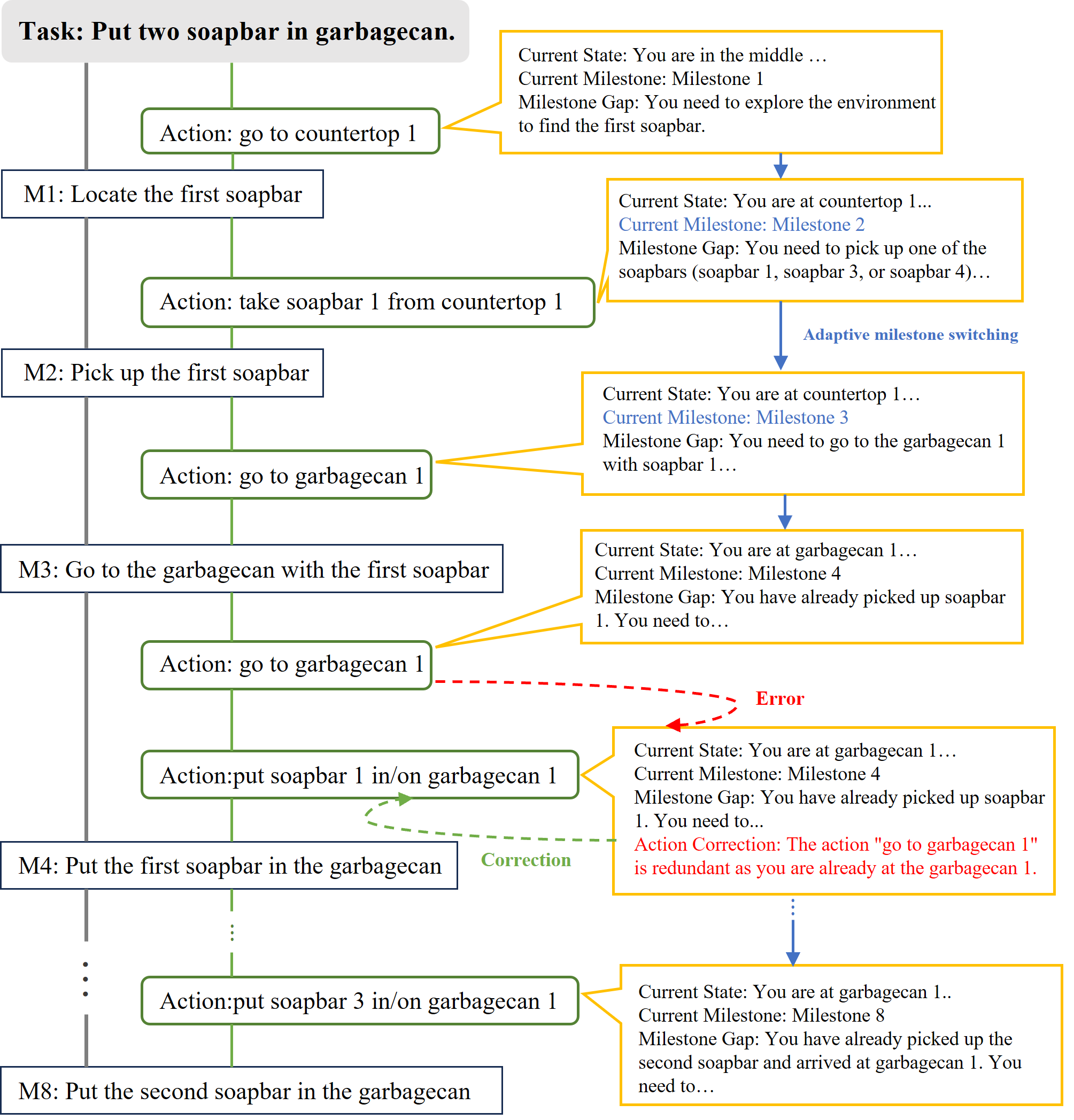}
    \caption{Illustration of \textsc{HiPlan}'s hierarchical guidance in the ``put two soapbar in garbagecan'' task. The diagram highlights milestone transitions (M), gap narrowing, and error correction via adaptive hints.}
    \label{fig:case_study}
\end{figure}

As shown in Fig.~\ref{fig:case_study}, \textsc{HiPlan} initially generates a global \textit{Milestone Action Guide} that decomposes the task into eight sequential and interconnected subgoals, spanning from locating the first soapbar through to disposing of the second one.

During execution, \textsc{HiPlan} leverages local \textit{Step-Wise Hints} to guide the agent through each subgoal. The hints are dynamically generated by LLM from prior milestone-level experience, enabling the agent to:
\begin{itemize}
    \item \textbf{Adaptively switch milestones} when preceding milestones are achieved, directing attention to the subsequent milestone (e.g., shifting the objective from ``Pick up soapbar'' to ``Go to garbagecan'' after an item is acquired).
    \item \textbf{Narrow milestone gaps} by retrieving action sequences from similar past milestones as references and analyzes the gap between the agent's current state and the milestone's objective to infer the most plausible next action.
    \item \textbf{Recall relevant memory} by injecting contextual information into the hint. For long-horizon tasks, this allows the agent to reuse prior knowledge established in earlier steps (e.g., re-visiting a known soapbar location), thus maintaining context and avoiding redundant actions.
    \item \textbf{Correct errors} by detecting deviations from expected progress (e.g., attempting to drop an item before reaching the garbagecan) and injecting targeted corrections in the hint.
\end{itemize}

Compared to flat structure baselines like REACT or TRAD, which may lack phased goal guidance and step-level error-awareness, \textsc{HiPlan} achieves more coherent and robust progression through long-horizon tasks.

\section{Conclusion}

In this paper, we introduced \textsc{HiPlan}, a hierarchical planning framework that enhances LLM-based agents' ability to tackle complex, long-horizon tasks by synergistically combining global milestone action guides with adaptive step-wise hints. This integration addresses critical challenges in maintaining global coherence while adapting actions to dynamic local contexts, leveraging milestone-level experience reuse to balance generalization and specificity effectively. Our extensive experiments demonstrate that \textsc{HiPlan} significantly improves success rates and robustness across diverse benchmarks and model architectures, outperforming strong baselines.

While \textsc{HiPlan} marks a substantial advancement, several avenues remain for future exploration. One promising direction is extending the framework to a broader range of tasks and domains to assess its generalizability and scalability. Additionally, we plan to investigate methods for summarizing and abstracting experience gained through step-wise hints, enabling effective cross-task knowledge transfer and improving adaptation in novel scenarios.

We believe that \textsc{HiPlan} represents a meaningful step toward empowering LLM-based agents with robust hierarchical reasoning and adaptive planning capabilities. By bridging global guidance and fine-grained adaptability, it lays the groundwork for more scalable, flexible, and intelligent autonomous systems capable of operating effectively in complex and dynamic real-world environments.

\bibliography{aaai2026}



\clearpage

\appendix

\section{Further Analysis: Step Efficiency}

In addition to task success rates, a critical measure of an agent's planning capability is its efficiency, i.e., the ability to complete tasks in a minimal number of steps. Inefficient planning often leads to redundant actions, error-prone exploration, and failure to complete tasks within a limited step-horizon. 

\begin{figure}[t]
    \centering
\includegraphics[width=\columnwidth]{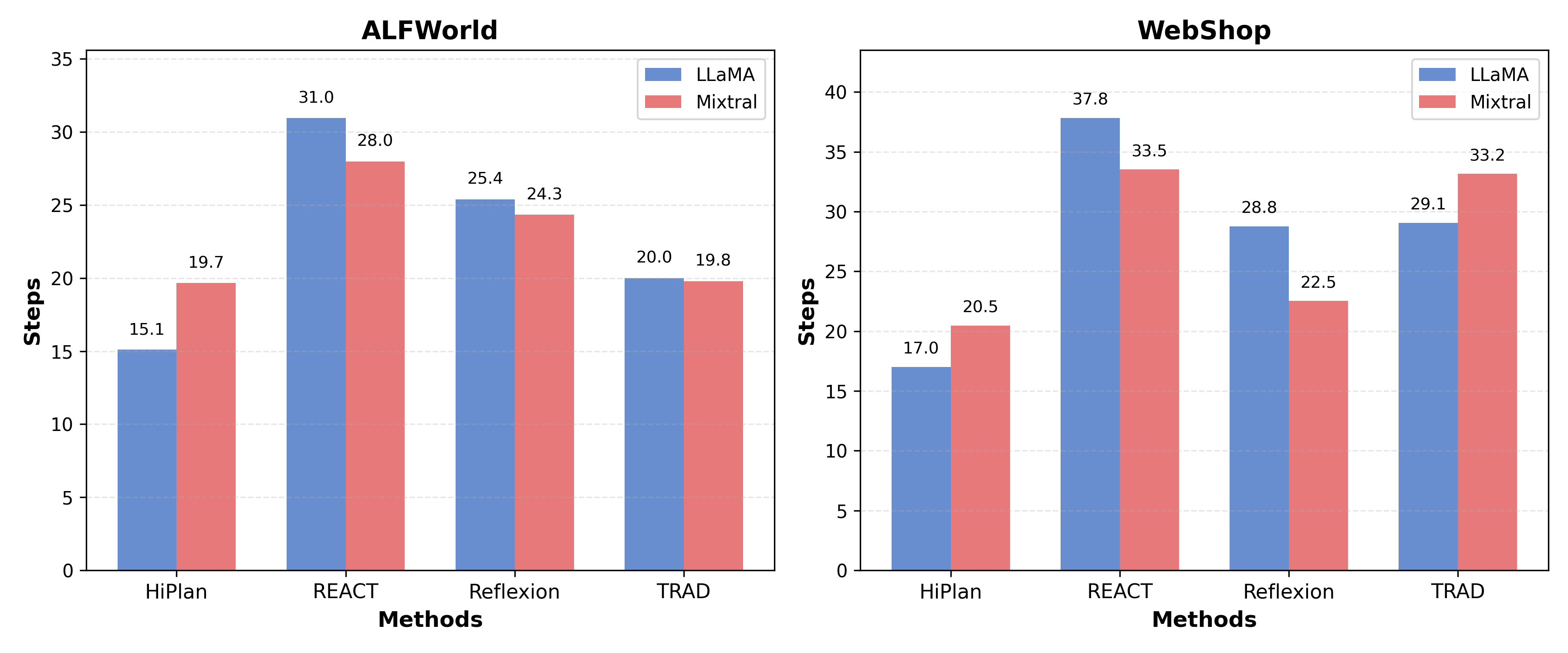}
    \caption{Average steps to task completion on ALFWorld and WebShop. \textsc{HiPlan} consistently achieves goals in fewer steps than baseline methods across both benchmarks, highlighting its superior planning efficiency.}
    \label{fig:step_analysis}
\end{figure}

Figure \ref{fig:step_analysis} illustrates the average number of steps taken by different methods across the ALFWorld and WebShop benchmarks. \textsc{HiPlan} consistently requires significantly fewer steps to complete tasks compared to all baselines. On average, \textsc{HiPlan} achieves a remarkable reduction of 28\% in steps on ALFWorld and 37\% on WebShop.

\paragraph{ALFWorld}
\textsc{HiPlan} outperforms all baselines by a significant margin, reducing the average number of steps by up to 51\% compared to REACT and by up to 41\% compared to Reflexion. Even against the strong retrieval-based TRAD baseline, \textsc{HiPlan} achieves a 25\% reduction in required steps. The improvements hold consistently for both LLaMA and Mixtral backbones.

\paragraph{WebShop}
A similar trend is observed on WebShop. \textsc{HiPlan} reduces the planning steps by up to 55\% compared to REACT and by over 40\% relative to Reflexion and TRAD.

\paragraph{Analysis}
We attribute the superior step efficiency of \textsc{HiPlan} to its hierarchical guidance mechanism: global milestone action guides maintain strategic direction, while adaptive step-wise hints correct deviations and eliminate redundant exploration. This dual-level guidance enables agents to advance towards goals in a more focused and error-averse manner, leading to not only higher success rates but also markedly shorter trajectories. 
In contrast, REACT often exhibits meandering behavior due to its reactive nature, and Reflexion is prone to repeated similar mistakes as it relies on post-hoc, episode-level feedback after a failure.
While TRAD shows competitive performance on task success rate, it still requires more steps to complete tasks than \textsc{HiPlan}, suggesting that its low-level trajectory demonstrations may guide agents along suboptimal planning paths.

The substantial drop in completion steps across heterogeneous tasks and LLMs confirms the generalizability and effectiveness of \textsc{HiPlan}'s design for long-horizon, complex decision-making tasks.

\section{Milestone Library Construction Details}

The milestone library construction process relies on an automated pipeline that extracts meaningful subgoals from expert demonstration trajectories. We employ GPT-4o to segment each trajectory into semantically coherent milestones using the prompt shown in Figure~\ref{fig:prompt_milestone_extraction}.

\subsubsection{Milestones Extraction and Trajectory Segmentation}

The extraction process operates on successful task demonstrations $\mathcal{D} = \{(\tau^{(i)}, \xi^{(i)})\}_{i=1}^N$, where each trajectory $\xi^{(i)} = \{(o_1^{(i)}, a_1^{(i)}), ..., (o_{T^{(i)}}^{(i)}, a_{T^{(i)}}^{(i)})\}$ corresponds to task $\tau^{(i)}$. For each trajectory, the LLM identifies key milestones and maps trajectory segments to corresponding milestones. 

For each identified milestone $k \in \{1, \ldots, K^{(i)}\}$, the LLM provides:
\begin{itemize}
    \item A description $m_k^{(i)}$ capturing the subgoal
    \item Action indices $\mathcal{I}_k^{(i)} = \{t_{start}^{(k)}, \ldots, t_{end}^{(k)}\}$ corresponding to the trajectory segment
\end{itemize}

The corresponding trajectory segment is then extracted as:
\[
\zeta_k^{(i)} = \{(o_t^{(i)}, a_t^{(i)}) : t \in \mathcal{I}_k^{(i)}\}
\]

This segmentation ensures that each $\zeta_k^{(i)}$ represents a cohesive sub-trajectory that accomplishes a specific intermediate objective toward the overall task completion.

\subsubsection{Indexing and Storage}

The constructed milestone library maintains a hierarchical organization with two-level indexing. At the task level, entries are indexed by task embedding vectors $v_\text{task}^{(i)}$ for efficient task-level retrieval. At the milestone level, individual milestones are indexed by their semantic embeddings $v_\text{milestone}^{(i,k)}$, enabling fine-grained milestone-level retrieval. Each entry stores tuples of the form $(v_\text{task}^{(i)}, v_\text{milestone}^{(i,k)}, m_k^{(i)}, \zeta_k^{(i)})$. All embeddings are computed using the SentenceTransformers model \texttt{all-mpnet-base-v2}.

The detailed statistics of the constructed milestone library are summarized in Table~\ref{tab:milestone_library_stats}.

\begin{table*}[h]
\centering
\begin{tabular}{l|cc|c}
\toprule
\textbf{Metric} & \textbf{ALFWorld} & \textbf{WebShop} & \textbf{Total} \\
\midrule
Source demonstrations & 500 & 77 & 577 \\
Avg. milestones per trajectory & 5.89 & 5.00 & 5.77 \\
Avg. actions per milestone & 1.78 & 1.11 & 1.70 \\
Total entries & 2,945 & 385 & 3,330 \\
\bottomrule
\end{tabular}
\caption{Statistics of the constructed milestone library across ALFWorld and WebShop datasets.}
\label{tab:milestone_library_stats}
\end{table*}

\section{Experience Retrieval Implementation}

This section provides detailed implementation of the experience retrieval mechanisms used in \textsc{HiPlan} for generating global milestone action guides and step-wise hints.

\subsection{Task-Level Retrieval for Global Guidance}

For generating the milestone action guide $G_\tau$, \textsc{HiPlan} retrieves $M=2$ most similar tasks from the milestone library. The retrieval process follows these steps:

\textbf{Task Embedding and Similarity Computation:} Given a test-time task instruction $\tau$, we first encode it into a dense vector using SentenceTransformers (\texttt{all-mpnet-base-v2}):
\begin{align}
v_\tau = \text{Embed}(\tau)
\end{align}

The similarity between the query task and each stored task is computed using normalized dot product:
\begin{align}
\text{sim}(v_\tau, v_{\text{task}^{(i)}}) = \frac{v_\tau \cdot v_{\text{task}^{(i)}}}{\|v_\tau\| \|v_{\text{task}^{(i)}}\|}
\end{align}

\textbf{Ranking and Selection:} Tasks are ranked by similarity in descending order, and the top-$M=2$ most similar tasks are selected:
\begin{align}
\{(\tau^{(j)}, \xi^{(j)}, G_{\tau^{(j)}})\}_{j=1}^M = \text{TopK}(\text{sim}(v_\tau, v_{\text{task}^{(i)}}), M)
\end{align}

\textbf{Length-based Re-ranking:} The retrieved demonstrations are further re-ranked by trajectory length to prioritize shorter, more generalizable examples that provide clearer structural guidance.

\subsection{Milestone-Level Retrieval for Local Guidance}

For generating step-wise hints $h_t$, \textsc{HiPlan} retrieves top-$P=2$ most similar milestones from the milestone library. The process follows similar embedding and similarity computation steps as global guidance retrieval, but operates at the milestone-level rather than task-level.

\textbf{Trajectory-level Deduplication:} Since similar milestones may exist within the same trajectory and could be recalled together, we enforce that no two selected milestones come from the same expert trajectory to ensure the diversity of demonstrations. This prevents over-reliance on specific trajectories and enhances generalization.

\textbf{Trajectory Segment Recovery:} For each retrieved milestone, we access the corresponding pre-stored trajectory segment that completes the milestone. For each trajectory segment, we include one additional action step to provide forward-looking context.

We set $M=2$ for global guidance retrieval and $P=2$ for step-wise hint retrieval based on empirical observations. These values align with established practices in in-context learning, where 2-3 examples typically yield optimal performance without introducing excessive noise or computational overhead.

\section{Error Mode Analysis}

Despite the overall strong performance of \textsc{HiPlan} across diverse long-horizon tasks, we observe certain failure cases that highlight challenges in reasoning under partial observability and language ambiguity. In this section, we analyze the common error modes on the ALFWorld and WebShop benchmarks. These cases help reveal not only the complexity of the task environments but also opportunities for further enhancing hierarchical planning and adaptive reasoning.

\subsection{ALFWorld}

ALFWorld poses embodied household tasks in a simulated environment with partially observable textual descriptions. We identify the following types of failure patterns:

\paragraph{Semantic Misidentification of Task Entities.}
In some tasks, the agent misidentifies the correct object due to persistent failure in locating the instructed item, ultimately defaulting to a semantically or functionally similar alternative. For instance, when tasked with retrieving an ``apple,'' the agent may eventually pick up a ``tomato'' after repeated unsuccessful attempts to find the target. This behavior highlights a core challenge in balancing flexibility and strict instruction adherence: while the agent adapts to the environment by choosing an available, plausible object, it risks deviating from the intended goal when precise specification is essential.

\paragraph{Inefficient Exploration and Repetition.}
We observe cases where the agent repeatedly visits previously checked locations, especially in environments with many distractors. For example, in the task requiring the agent to ``heat an egg and put it in the garbagecan,'' over 40 steps were spent re-opening the same refrigerator, drawers, and non-existent objects like ``diningtable 1'' before finally locating the egg in an unexpected container (the garbagecan itself). Such inefficient loops can be attributed to the partial observability of the environment and the difficulty in reasoning about uncommon object placements. While \textsc{HiPlan} provides milestone-level direction, it occasionally lacks fine-grained memory consolidation to prevent redundant behavior in exploration-heavy subtasks.

\paragraph{Failure to Satisfy Implicit Task Constraints.}
In certain cases, the agent transitions to the next milestone without fully completing the current one, due to an inaccurate judgment of milestone completion based on recent actions and observations. For example, in the task ``put a cool tomato in the microwave,'' the agent moves the tomato into the microwave without successfully performing the cooling step—despite this being an explicitly defined milestone in the generated action guide. This indicates that while \textsc{HiPlan} maintains a structured global plan, it occasionally struggles to determine whether a milestone has been genuinely achieved from textual cues alone. Such errors reflect the inherent ambiguity in inferring milestone completion in text-only, partially observable environments, where changes in object state (e.g., cooled vs. not cooled) may not be explicitly described.

\subsection{WebShop}

WebShop introduces a distinct class of long-horizon reasoning tasks focused on goal-directed online shopping. We observe the following error modes:

\paragraph{Specification Drift in Product Selection.}
Some agents fail to strictly adhere to all constraints specified in the instruction, such as size, material, or color. In one example, although the user explicitly requested a product with ``6.76 fl oz'' volume, the agent skipped the option selection step and directly purchased the default variant. This behavior stems from partial satisfaction of constraints (e.g., product match but wrong size) and suggests limitations in multi-attribute reasoning under ambiguous product descriptions.

\paragraph{Action Loops and Invalid Sequences.}
We observe cases where the agent repeatedly issues invalid or ineffective actions, often resulting in prolonged sequences of failure. These loops typically emerge when the agent attempts to execute actions that are not applicable to the current page or context—such as clicking a button that does not exist on the current interface. This suggests that, in certain situations, the agent fails to properly analyze the local environment before acting and does not robustly retry alternative strategies when actions are invalid. Such behavior underscores the challenge of accurately grounding decision-making in transient UI states within text-based environments, and the occasional lack of sufficient corrective mechanisms in the step-wise hint generation process.

\paragraph{Underutilization of Available Information.}
In certain tasks, agents overlook available high-quality candidates presented early in the session. For example, a correct item appeared in the first search results but was ignored in favor of less suitable alternatives. In some cases, agents also skipped detailed product descriptions and features, leading to premature purchase decisions that violated key constraints like ``lace closure'' or ``water resistance.'' These failures indicate occasional overemphasis on surface cues (e.g., color or title keywords) over deeper semantic validation.

\section{Baselines Implementation Details}
\subsection{REACT}
We used the official REACT codebase, which includes experiments on both ALFWorld and WebShop. All settings were aligned with ours. For ALFWorld, we used the same 134 out-of-distribution tasks with a maximum of 50 steps. For WebShop, we used the same 200 test queries with a 40-step limit.

\subsection{Reflexion}
We adopted the official RefleXion codebase, which supports experiments on ALFWorld and WebShop. All settings were adjusted to match ours. On ALFWorld, we used the same 134 out-of-distribution tasks with a 50-step limit, running three reflection rounds. On WebShop, we used the same 200 test queries with a 40-step cap, also running three rounds.

\subsection{Trad}
For the ALFWorld dataset, we adopted the original TRAD implementation, which includes experiments on this benchmark. We reused the official codebase and modified all settings to match those used by our method. Specifically, we used the same 134 out-of-distribution tasks and capped the maximum number of steps at 50.

For the WebShop dataset, the original TRAD paper did not provide experimental results. Therefore, we carefully implemented the method based on the official description. All prompts used in our implementation are shown in Fig.~\ref{fig:prompt_extract_thought_trad} and Fig.~\ref{fig:prompt_action_trad}. During the WebShop experiments, we used the same 77 expert demonstrations as our method for retrieval and the same 200 test tasks. The maximum number of allowed steps was set to 40.

\begin{figure*}[t]
    \centering
    \small
    \begin{tcolorbox}[colframe=black, colback=white, width=\textwidth, boxrule=0.2mm]
You are given a household manipulation task and its ideal action trajectory. Your job is to:
\\
1 Identify the key milestones (subgoals or logical steps) necessary to complete the task.
\\
2 Divide the action trajectory into segments, where each segment corresponds to a milestone.
\\
3 For each milestone, list the indices of the actions (from the trajectory) that belong to that milestone.
\\ \hspace*{\fill} \\
Instructions:
\\
- Only consider the actions (ignore the observations) when mapping actions to milestones.
\\
- Each milestone should represent a clear, meaningful subgoal within the overall task.
\\
- The output should be a JSON array, where each object contains:
\\
    - ``milestone'': a concise description of the subgoal.
\\
    - ``actions'': a list of action indices.
\\ \hspace*{\fill} \\
Example 1:
\\
Input:
\\
- Task: put a egg in microwave.
\\
- Trajectory: TRAJECTORY EXAMPLE1
\\
Output: MILESTONES EXAMPLE1
\\ \hspace*{\fill} \\
Example 2:
\\
Input:
\\
- Task: put a clean soapbar in countertop.
\\
- Trajectory: TRAJECTORY EXAMPLE2
\\
Output: MILESTONES EXAMPLE2
\\ \hspace*{\fill} \\
Input:
\\
Task: \{TASK\}
\\
Trajectory:
\\
\{TRAJECTORY\}
\\ \hspace*{\fill} \\
Now, please generate the milestone list and map the actions to each milestone in the same format:
    \end{tcolorbox}
    \caption{Prompt for milestone extraction and trajectory segmentation.}
    \label{fig:prompt_milestone_extraction}
\end{figure*}

\begin{figure*}[t]
    \centering
    \small
    \begin{tcolorbox}[colframe=black, colback=white, width=\textwidth, boxrule=0.2mm]
   Generate a thoughtful reasoning for the following shopping action.
\\ \hspace*{\fill} \\
Shopping Task: TASK

Current Trajectories: TRAJECTORIES

Next Action: NEXT ACTION

Examples: EXAMPLES
\\ \hspace*{\fill} \\
Based on the observation, what is the reasoning for taking this action? Express this as a first-person thought that explains the strategic thinking.
\\ \hspace*{\fill} \\
Thought:
    \end{tcolorbox}
    \caption{Prompt for generating thoughts on expert demonstrations in TRAD method.}
    \label{fig:prompt_extract_thought_trad}
\end{figure*}

\begin{figure*}[t]
    \centering
    \small
    \begin{tcolorbox}[colframe=black, colback=white, width=\textwidth, boxrule=0.2mm]
   You are a helpful assistant to do online shopping.
   
- You will be given an instruction about what to buy.

- You need to navigate on a website to purchase an item that meets the instruction.

- You can see the web page content and your task is to output the next action.

- The actions must be in the format of `search[keywords]' or `click[button]'.

- You should pay attention to the buttons available in the observation to decide your next action.

- For example, if you want to search again, you may need to `click[Back to Search]' first.
\\ \hspace*{\fill} \\
Here are examples to help with this shopping task:
DEMONSTRATIONS

Now for the current task:
TRAJECTORIES
    \end{tcolorbox}
    \caption{Prompt for taking the next action or generating thought in TRAD method.}
    \label{fig:prompt_action_trad}
\end{figure*}

\section{Prompts used in \textsc{HiPlan}}

\subsection{ALFWorld}

We present all prompts utilized in the \textsc{HiPlan} framework on ALFWorld in the following figures: Fig.~\ref{fig:prompt_milestone_alfworld}, Fig.~\ref{fig:prompt_step_wise_hint_alfworld}, Fig.~\ref{fig:prompt_action_alfworld}.

\begin{figure*}[t]
    \centering
    \small
    \begin{tcolorbox}[colframe=black, colback=white, width=\textwidth, boxrule=0.2mm]
    You are a professional planner specializing in breaking down complex tasks into clear, milestone-driven action guides based on expert examples.
\\ \hspace*{\fill} \\
Your instructions:

- Carefully study the provided example(s) and reproduce their style exactly in your answer.

- Match the examples in wording, logic, and step order as closely as possible.

- Do not add, remove, or rephrase steps unless strictly necessary to fit the new task.

- Organize your solution as a sequence of major milestones, each representing a key stage in accomplishing the task, just as in the examples.

- Each milestone should be concise and actionable, using the same pattern and phrasing style as the examples.
\\ \hspace*{\fill} \\
Example:
EXAMPLES

Task: 
TASK
\\ \hspace*{\fill} \\
Following the provided style and format, outline a milestone-based action guide for the given task (no unnecessary explanations).
Milestone action guide:
    \end{tcolorbox}
    \caption{Prompt for LLM in generating milestone action guide on ALFWorld based on milestone library.}
    \label{fig:prompt_milestone_alfworld}
\end{figure*}

\begin{figure*}[t]
    \centering
    \small
    \begin{tcolorbox}[colframe=black, colback=white, width=\textwidth, boxrule=0.2mm]
    You are an assistant guiding an agent that performs household tasks in a simulated environment. Your task is to generate a step-wise hint that helps the agent for each action step.
\\ \hspace*{\fill} \\
Your hint must be based on the following:

- Current Task: The overall task objective.

- Current Trajectory: The sequence of actions and observations up to the current step.

- Milestone-Based Guide: A list of milestones required to complete the task.

- Similar Trajectories: Retrieved segments of successful past action/observation trajectories matching the current task and state, including those aligned with the most similar milestone. 
\\ \hspace*{\fill} \\
Analyze the agent's current state, the milestone-based guide, and similar trajectories, and generate the step-wise hint including only the following fields:

- Current State: Briefly describe the agent’s relationship to the environment and relevant objects. 

- Current Milestone: Specify which milestone in the plan the agent should currently be addressing.

- Milestone Gap: State what remains to be done to complete the current milestone. 

- Action Correction (optional): Include this field only if the recent action is incorrect or deviates from the milestone. In that case, point out the error and specify the correct direction or action to take. If the agent’s recent action and state are correct, omit this field.
\\ \hspace*{\fill} \\
Guidelines:

- Be concise and specific.

- Always relate feedback to the current milestone; once a milestone is complete, advance to the next.

- If the agent has made a mistake , clearly indicate the error using the Action Correction field; otherwise, do not include this field.
\\ \hspace*{\fill} \\
Format your output as follows (omit Action Correction if not needed):

Current State: [brief description of agent’s relevant context]

Current Milestone: [The current goal that needs to be achieved. Milestone X – description]

Milestone Gap: [what still needs to be done, grounded only in observed information and milestones]

Action Correction: [only if an explicit correction is needed; otherwise omit this field]
\\ \hspace*{\fill} \\
Here are two examples:
[TWO FIXED EXAMPLES FOR STEP-WISE HINTS GENERATION]
\\ \hspace*{\fill} \\
Your Input:  
Current Task: 
TASK

Current Trajectory: 
TRAJECTORIES

Milestone-Based Guide: 
MILESTONE ACTION GUIDE

Similar Trajectories: 
MILESTONE-LEVEL DEMONSTRATIONS 
\\ \hspace*{\fill} \\
Now, please generate the hint for the next action (no unnecessary explanations).

Output:
    \end{tcolorbox}
    \caption{Prompt for generating step-wise hints on ALFWorld.}
    \label{fig:prompt_step_wise_hint_alfworld}
\end{figure*}

\begin{figure*}[t]
    \centering
    \small
    \begin{tcolorbox}[colframe=black, colback=white, width=\textwidth, boxrule=0.2mm]
    Task: Interact with a household to complete tasks involving placing/operating on object(s) to/in/on a target, and wait for next observation.
\\ \hspace*{\fill} \\
Action Space:

1. Go to [target]: Move to the target; observe its contents or state (opened/closed).

2. Open [target]: Open closable targets. Observe contents. Only cabinets, drawers, fridges, safes, and microwaves can be opened.

3. Take [object] from [target]: Pick up one object from the target. You can only take one object at the same time.

4. Put [object] in/on [target]: Place held object on/in target (must be at target).

5. Clean [object] with [target]: Clean an object at the sinkbasin after moving there. Other items in/on the sinkbasin don't affect cleaning.

6. Heat [object] with [target]: Heat an object in the microwave after moving there. Other items inside don't affect heating.

7. Cool [object] with [target]: Cool an object in the fridge after moving there, regardless of other items inside.

8. Use [object]: The object should be a desklamp. Use the desklamp where it is.
\\ \hspace*{\fill} \\
You can refer to the following milestone-based action guide proposed for this task to take action:
MILESTONE ACTION GUIDE

Here are two examples:
TASK-LEVEL DEMONSTRATIONS

Your task and trajectories are as follows:
TRAJECTORIES

You can follow the hint to take the next action:
STEP-WISE HINT
\\ \hspace*{\fill} \\
Now, take the next action for your task (no unnecessary explanations):
    \end{tcolorbox}
    \caption{Prompt for agent to take the next action on ALFWorld.}
    \label{fig:prompt_action_alfworld}
\end{figure*}

\subsection{WebShop}
We present all prompts utilized in the \textsc{HiPlan} framework on ALFWorld in the following figures: Fig.~\ref{fig:prompt_step_wise_hint_webshop}, Fig.~\ref{fig:prompt_action_webshop}. And the milestone extraction method is identical to that in ALFWorld.

\begin{figure*}[t]
    \centering
    \small
    \begin{tcolorbox}[colframe=black, colback=white, width=\textwidth, boxrule=0.2mm]
    You are an assistant guiding an agent that performs online shopping tasks in a simulated webshop environment. Your task is to generate a step-wise hint that helps the agent complete the current milestone.
\\ \hspace*{\fill} \\
WebShop tasks follow 5 milestones:
1. Search phase: Formulating and executing search query

2. Browse phase: Examining search results and selecting items  

3. Evaluation phase: Assessing product details

4. Configuration phase: Selecting product options

5. Purchase phase: Completing the transaction

IMPORTANT CONSTRAINTS:
- The agent CANNOT click [Next] to view more search results. They must find a suitable product on the first page.

- Pay VERY CLOSE ATTENTION to the current observation - only suggest actions that are actually available in the current state.

- CAREFULLY ANALYZE THE ENTIRE TRAJECTORY HISTORY to avoid suggesting actions that have already been taken.
\\ \hspace*{\fill} \\
Analyze the agent's current state, the milestone-based guide, and similar trajectories, and generate the step-wise hint including only the following fields:

- Current State: VERY PRECISELY identify the exact page type the agent is currently on.

- Current Milestone: Specify which of the milestones the agent should currently be addressing.  WebShop milestones can be addressed non-sequentially if needed. 

- Milestone Gap: First analyze what still needs to be done to complete the current milestone (the gap). Then, based on this analysis, recommend ONLY the NEXT single action in exact format. 

- Action Correction (optional): Include this field ONLY if the MOST RECENT action resulted in ``Invalid action!'' error. Point out exactly why the action was invalid and specify the correct action to take. OMIT THIS FIELD ENTIRELY if the most recent action was valid.
\\ \hspace*{\fill} \\
Format your output as follows (omit Action Correction if not needed):

Current State: [describe exact page type]

Current Milestone: [Milestone X – description]

Milestone Gap: [Analysis of what's needed to complete milestone + ONLY the next immediate action in exact format]

Action Correction: [only if the MOST RECENT action resulted in ``Invalid action!'']
\\ \hspace*{\fill} \\
Here are several examples:
[FOUR FIXED EXAMPLES FOR STEP-WISE HINTS GENERATION]
\\ \hspace*{\fill} \\
Your Input:  
Current Task: 
TASK

Current Trajectory: 
TRAJECTORIES

Similar Trajectories: 
MILESTONE-LEVEL DEMONSTRATIONS 
\\ \hspace*{\fill} \\
Now, please generate the hint for the next action:

    \end{tcolorbox}
    \caption{Prompt for generating step-wise hints on WebShop.}
    \label{fig:prompt_step_wise_hint_webshop}
\end{figure*}

\begin{figure*}[t]
    \centering
    \small
    \begin{tcolorbox}[colframe=black, colback=white, width=\textwidth, boxrule=0.2mm]
    Task: Complete online shopping tasks by searching for products, navigating product pages, selecting appropriate options, and making purchases.
\\ \hspace*{\fill} \\
Action Space:

1. search[query]: Search for products using specific keywords related to the shopping requirement.

2. click[button/option]: Click on buttons, product links, or options.
   \\ \hspace*{\fill} \\
IMPORTANT CONSTRAINTS:

- You CANNOT click [Next] to view more search results. You must find an appropriate product on the first page.

- Focus on products that most closely match the requirements in the task.
\\ \hspace*{\fill} \\
You can refer to the following milestone-based action guide proposed for this task:
MILESTONE ACTION GUIDE

Here are examples of similar shopping tasks:
TASK-LEVEL DEMONSTRATIONS

Your current shopping task and actions taken so far:
TRAJECTORIES

You can follow the hint for the next action:
STEP-WISE HINT
\\ \hspace*{\fill} \\
Now, take the next action for your shopping task. Provide only the action in the format specified above:

Action: 
    \end{tcolorbox}
    \caption{Prompt for agent to take the next action on WebShop.}
    \label{fig:prompt_action_webshop}
\end{figure*}

\end{document}